\documentclass[10pt,twocolumn,letterpaper]{article}

\usepackage{wacv}
\usepackage{times}
\usepackage{epsfig}
\usepackage{graphicx}
\usepackage{amsmath}
\usepackage{amssymb}
\usepackage{booktabs}
\usepackage{multirow, makecell}
\usepackage{bm}
\usepackage{subcaption}
\usepackage{csquotes}
\usepackage{algorithm}
\usepackage{algpseudocode}
\usepackage[accsupp]{axessibility}

\newcommand{\sota}{state-of-the-art }
\newcommand{\Sota}{State-of-the-art }
\DeclareMathOperator*{\argmin}{arg\,min}

\newcommand{\partatwo}{Part-$A^2$}
\newcommand{\bestresult}[1]{\text{\textbf{#1}}}
\newcommand{\secbresult}[1]{\underline{#1}}

\graphicspath{ {./images/} }

\def\wacvPaperID{998} 

\wacvfinalcopy 

\ifwacvfinal
\usepackage[breaklinks=true,bookmarks=false]{hyperref}
\else
\usepackage[pagebackref=true,breaklinks=true,colorlinks,bookmarks=false]{hyperref}
\fi

\begin{document}

\title{
  SAILOR: Scaling Anchors via Insights into Latent Object Representation
}

\author{Du\v{s}an Mali\'c$^{1, 2}$
\and
Christian Fruhwirth-Reisinger$^{1, 2}$
 \and
 Horst Possegger$^{1}$
 \and
 Horst Bischof$^{1, 2}$
 \and \\
 $^1$Institute of Computer Graphics and Vision, Graz University of Technology \\
 $^2$Christian Doppler Laboratory for Embedded Machine Learning
\and
{\tt\small \{dusan.malic, christian.reisinger, possegger, bischof\}@icg.tugraz.at}
}


\maketitle
\thispagestyle{empty}

\begin{abstract}
  LiDAR 3D object detection models are inevitably biased towards their training dataset.
  The detector clearly exhibits this bias when employed on a target dataset, particularly towards object sizes.
  However, object sizes vary heavily between domains due to, for instance, different labeling policies or geographical locations.
  \Sota unsupervised domain adaptation approaches outsource methods to overcome the object size bias.
  Mainstream size adaptation approaches exploit target domain statistics, contradicting the original unsupervised assumption.
  Our novel \emph{unsupervised anchor calibration} method addresses this limitation.
  Given a model trained on the source data, we estimate the optimal target anchors in a completely unsupervised manner.
  The main idea stems from an intuitive observation: by varying the anchor sizes for the target domain, we inevitably introduce noise or even remove valuable object cues.
  The latent object representation, perturbed by the anchor size, is closest to the learned source features only under the optimal target anchors.
  We leverage this observation for anchor size optimization.
  Our experimental results show that, without any retraining, we achieve competitive results even compared to \sota weakly-supervised size adaptation approaches.
  In addition, our anchor calibration can be combined with such existing methods, making them completely unsupervised.
\end{abstract}

\section{Introduction}
Acquiring and labeling data for the training of 3D object detectors requires considerable effort.
The sheer size and the underlying unorganized structure of point clouds make this process cumbersome.
Detecting objects in a point cloud can be difficult even for humans since an object may contain just a few points.
Moreover, during labeling, an expert will usually examine a single LiDAR frame from multiple viewpoints to account for unavoidable occlusions and truncations, usually switching the context between an image and a point cloud.
These interruptions substantially increase labeling time and, consecutively, labeling cost.
\begin{figure}[t]
  \begin{center}
    \includegraphics[width=\linewidth]{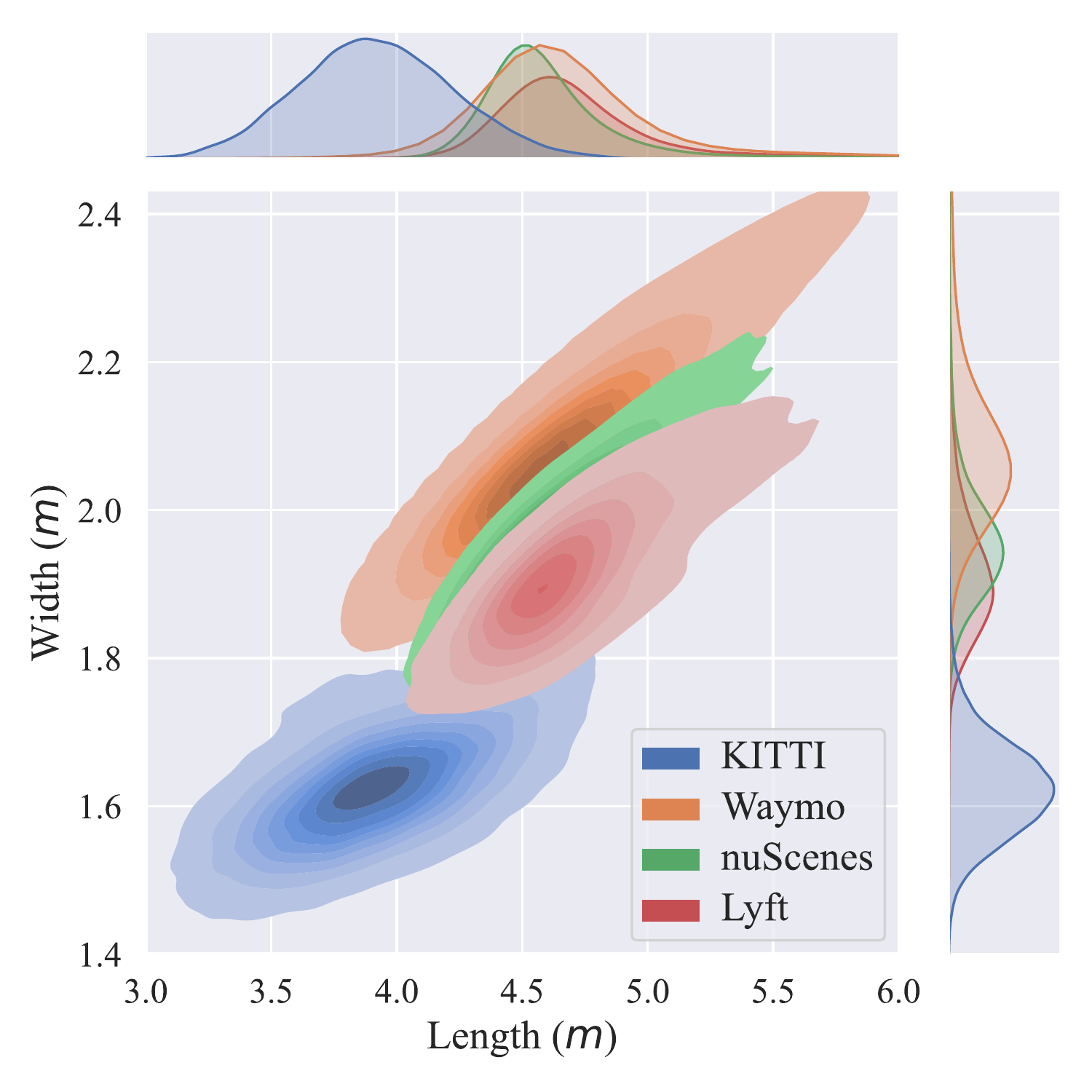}
  \end{center}
  \caption{
    Box sizes, showing the domain gap across datasets.
    Boxes are generated by a fully-supervised oracle model trained on KITTI~\cite{3ddataset:kitti},  Waymo~\cite{3ddataset:waymo}, nuScenes~\cite{3ddataset:nuscenes} and Lyft~\cite{3ddataset:lyft}, respectively.
  }
  \label{fig:density_width_length_oracle}
\end{figure}

To mitigate the high labeling cost, our research community is making immense advances in unsupervised 3D domain adaptation~\cite{3dda:sfuda,3dda:train_germany,3dda:st3d}.
\Sota approaches typically rely on some variant of self-training~\cite{3dda:st3d++,3dda:st3d}, input transformation~\cite{3dda:eye2eye,3dda:lidar_distilation,3dda:spg}, feature alignment~\cite{3dda:srdan} and/or tracking~\cite{3dda:fast3d,3dda:explot_playbacks}.
These approaches share a common problem, initially reported by Wang \etal~\cite{3dda:train_germany}: the discrepancy between object sizes, as depicted in Figure~\ref{fig:density_width_length_oracle}, introduces a tremendous domain gap.

In 2D object detection, object sizes vary depending on the distance of an object from a sensor.
This naturally occurring augmentation increases a detector's robustness towards size variation.
In 3D object detection, however, the size of an object does not correlate with the distance from the sensor.
Instead, it depends on where the dataset has been acquired, \eg on average, vehicles in the USA are larger than the vehicles in Europe~\cite{3dda:train_germany}, or on the labeling policy, \eg Waymo Open Dataset~\cite{3ddataset:waymo} includes side mirrors of cars in its annotations, whereas KITTI~\cite{3ddataset:kitti} does not.

Object detectors are commonly evaluated using the Mean Average Precision (mAP) metric, which further utilizes Intersection over Union (IoU) for ground truth matching.
In 2D detection, IoU is calculated given the area of the predicted and ground truth bounding box, whereas in 3D, it is derived using the volumes of the predicted and the ground truth cuboid.
Given the additional dimension, the overall influence of the incorrect size prediction leads to an exponential accuracy decrease.
This correlation makes 3D detectors extremely volatile to changes in the object sizes.

Statistical Normalization (SN)~\cite{3dda:train_germany} has become the standard approach for bridging the size gap. 
It attempts to shift the source data statistics to the target statistics through deliberate scaling of the source annotations as a training augmentation.
Random Object Scaling (ROS)~\cite{3dda:st3d} strives to overcome this size bias without directed scaling.
Instead, it substantially augments the ground truth boxes via a wider range of scales.
Nevertheless, both approaches exploit key target domain insights, which are usually not available in an unsupervised setting.
First, they employ anchor sizes, manually optimized for the target domain.
Second, they manually refine the magnitude of the augmentation scales.
Lastly, the stochastic nature of these approaches does not allow a deterministic checkpoint selection, but rather the best performing one.


We propose SAILOR, a novel \emph{unsupervised anchor calibration pipeline} which estimates optimal target anchors given a pretrained source model.
Our objective is to identify target anchors under which the feature similarity between a source and a target domain is maximized.
We first establish a reference feature database by gathering instance features from the source domain.
In the next phase, we iteratively perturb the anchors by a small amount, and analogously compute a target feature database. 
The fitness of the target feature database to the reference feature database provides a stochastic gradient, which we employ in a stochastic optimization method.
Our method achieves similar improvements as SN or ROS, but does not introduce additional model parameters, does not require retraining, and does not require any knowledge from the target domain at all.


Experimental results on the autonomous driving datasets KITTI~\cite{3ddataset:kitti}, Waymo~\cite{3ddataset:waymo}, nuScenes~\cite{3ddataset:nuscenes} and Lyft~\cite{3ddataset:lyft} indicate large performance gains with minimal effort.
We demonstrate that a simple exchange of the source anchors with our optimized ones, leads to large precision gains on the target domain.
Our method is competitive even with the popular weakly-supervised approaches for bridging the object size domain gap, but costs only a fraction of the computation time, since we do not require retraining.
Additionally, our optimized anchors can also be used as an unsupervised prior to extend weakly-supervised methods, turning them into fully unsupervised approaches.


\section{Related Work}

\paragraph{3D Object Detection}
\Sota 3D detectors consist of a Region Proposal Network (RPN) followed by a detection head.
RPN first abstracts an input point cloud with a feature extractor, which is commonly point-based~\cite{3dod:pointrcnn}, voxel-based~\cite{3dod:parta2,3dod:second} or a hybrid between these two~\cite{3dod:pvrcnn}.
Afterwards, at each discrete location of the abstracted point cloud, RPN predicts an objectness probability and regresses coarse bounding boxes.
The magnitude of this regression is either absolute~\cite{3dod:centernet} or relative to \emph{anchor} values~\cite{3dod:voxelrcnn,3dod:pointpillars,3dod:pvrcnn,3dod:parta2,3dod:second}.
The anchors reduce the network's regression space in such a way that the predictions become residuals from a matched anchor.
The anchor values are commonly derived from the training data and have to be manually selected with care.
The following detection head refines such coarse predictions into the final predictions.

\paragraph{Unsupervised 3D Domain Adaptation}
A 3D detection model trained on one dataset usually does not generalize well to new, unseen data.
The source and target dataset difference is commonly referred to as the \emph{domain gap}.
Unsupervised domain adaptation aims to reduce the gap without any additional annotations.

Style-transfer approaches lessen the gap in the input space by transforming the target data into source-like data.
Wei \etal~\cite{3dda:lidar_distilation} demonstrate the generalization capabilities of a model trained on dense data augmented by pseudo-sparse point clouds derived from the original dense data.
Dropping LiDAR beams scales up with the size of a point cloud since it is not a costly operation.
However, it is ambiguous to what extent the background transformation helps and if it justifies a more complex conversion.
Modifying just the foreground objects by adding new points~\cite{3dda:eye2eye,3dda:spg} or reorganizing the existing ones~\cite{3dda:3dvfield} saves computational resources with a similar effect.

The sequential acquisition of autonomous driving datasets allows exploiting temporal consistency for the adaptation.
Some self-training approaches~\cite{3dda:sfuda,3dda:auto4d,3dda:explot_playbacks} refine per-frame detection via Multi Object Tracking (MOT)~\cite{3dmot:weng} and are thus able to improve na\"ive retraining by reducing the false positive and negative rate.
Even without explicit tracking, aggregating several LiDAR frames into a common reference frame already leads to denser static objects.
Detecting such objects is a trivial task for \sota 3D detectors~\cite{3dod:focalsconv,3dod:pvrcnn,3dod:second}.
By propagating high-quality detections to all aggregated frames, \cite{3dda:temorallbling} generates pseudo labels for even the more complex (sparse and static) samples.
FAST3D~\cite{3dda:fast3d} further extends this idea to moving objects, by leveraging scene flow~\cite{3dflow:pointpwc} for aggregation.

Besides tracking and flow pitfalls, \eg association, domain gap, \etc, temporal information may not always be available.
Generating pseudo labels on a per-frame basis has also been beneficial for self-training~\cite{3dda:pl_3dod}.
Leveraging the gradual improvement on the presented pseudo labels, memory ensembles~\cite{3dda:st3d++,3dda:st3d} enhance the robustness by cataloging the predictions during training.
An auxiliary loss~\cite{3dda:3dcoco,3dda:srdan} can further reduce the inevitable noise induced by the pseudo labels.
Still, the self-training is going to amplify the noise.
This is a well-known side effect called \emph{inductive bias}, which is generally resolved with the student-teacher paradigm~\cite{3dda:uncertain_mt,3dda:mlc}.

The self-training approaches are not able to independently overcome the object size bias induced by the source data.
The extent of the object size gap immensely influences the performance of the final model.
Therefore, \sota self-training approaches explicitly utilize mechanisms to mitigate the cross-domain size mismatch.

\paragraph{Overcoming Object Size Bias}
Mainstream methods exploit target domain knowledge to overcome the object size bias.
Wang \etal~\cite{3dda:train_germany} propose employing the target domain statistics in two ways.
Output Transformation (OT) directly modifies a predicted bounding box size by adding a residual.
The difference between an average source and a target sample defines this residual.
Statistical Normalization (SN) scales source ground truth boxes and points inside during training.
Analogously to OT, the scaled intensity is derived from the source and target dataset statistics.
To avoid precise target domain statistics that are usually unavailable, Random Object Scaling (ROS)~\cite{3dda:st3d} implements heavy size augmentation on the source ground truth objects.
However, increasing the network's search space leads to a frequent size prediction mismatch.

Contrary to the weakly-supervised approaches, we propose a completely unsupervised method, which specifically optimizes a model's anchors for the target domain.
Moreover, our approach does not require retraining and thus costs just a fraction of computational time.
We achieve a similar effect as these weakly-supervised methods, since more suitable target anchors alleviate the model's extrapolation efforts and implicitly provide improved detections.

\section{Unsupervised Anchor Calibration}
\label{sec:method}
\begin{figure*}
  \begin{center}
    \includegraphics[width=0.9\linewidth]{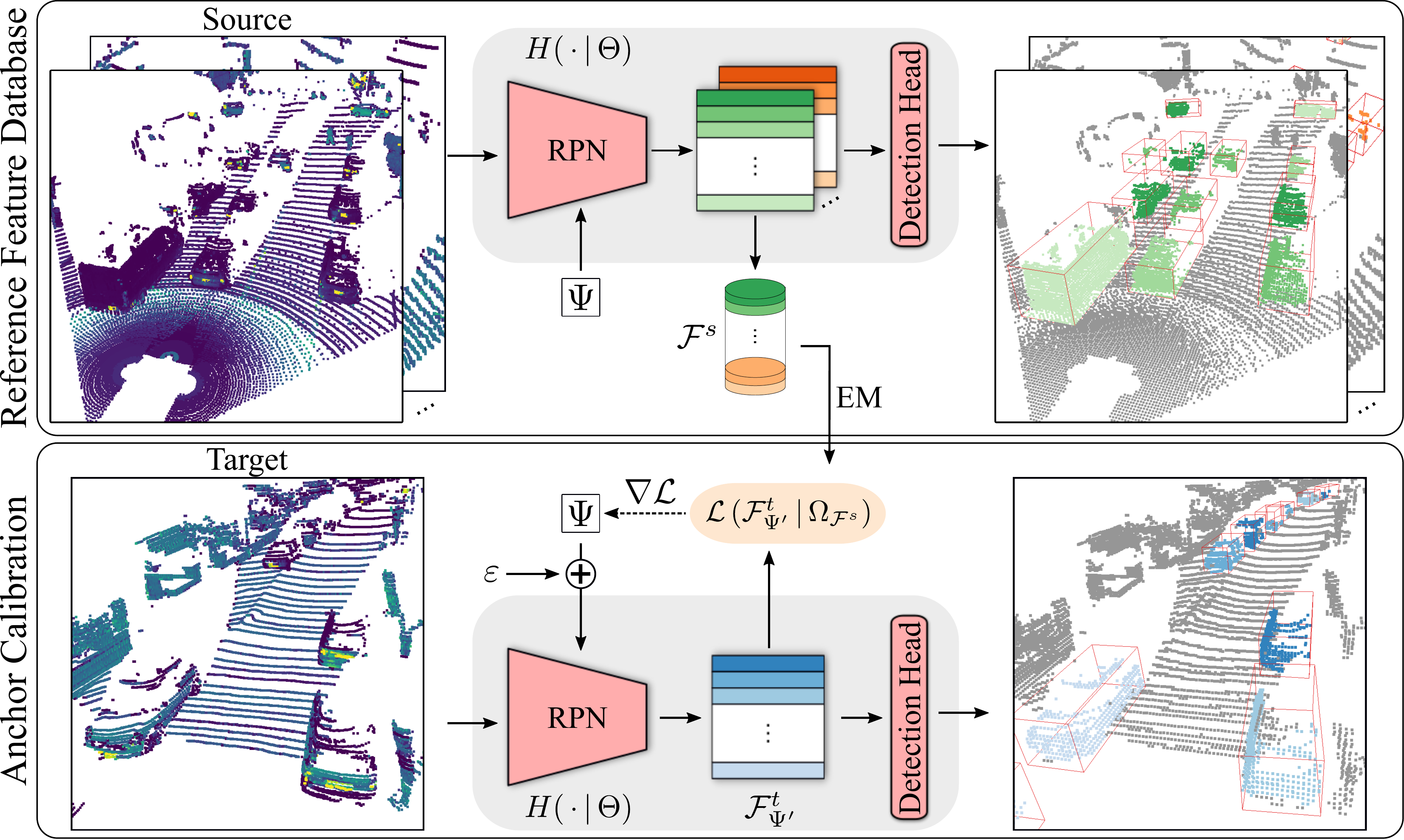}
  \end{center}
  \caption{
    Overview of SAILOR: Given a model $H ( \, \cdot \, | \, \Theta )$ with parameters $\Theta$ trained on the source data and anchors $\Psi \subset \Theta$ optimized for the same data, we first extract a source feature database $\mathcal{F}^s$ by accumulating Region Proposal Network's (RPN) predictions.
    We then fit a Gaussian Mixture Model (GMM) parameters $\Omega_{\mathcal{F}^s}$ to $\mathcal{F}^s$ using Expectation Maximization (EM)~\cite{stat:em}.
    During anchor calibration on the target domain, using the same model, we iteratively introduce small noise $\varepsilon$ to the source anchors and extract a temporary target feature database $\mathcal{F}^t_{\Psi'}$.
    We utilize a stochastic gradient $\nabla \mathcal{L}$, estimated from the fitness $\mathcal{L} (\mathcal{F}^t_{\Psi'} \, | \, \Omega_{\mathcal{F}^s})$, to directly adapt the source anchors to the target domain.
    We neither perform any further adaption, nor require any retraining.
  }
  \label{fig:method_overview}
\end{figure*}
We consider a 3D detection model $H( \, \cdot \, | \, \Theta ) = D \circ RPN$, with a Region Proposal Network $RPN$ and a detection head $D$.
Its parameters $\Theta$ are trained on a labeled source dataset $\mathcal{S} = \{(\boldsymbol{P}^s_i, \boldsymbol{Y}^s_i)\}_{i = 1}^{N_s}$, which contains LiDAR point clouds $\boldsymbol{P}^s$ and a respective set of labeled instances $\boldsymbol{Y}^s$.
The target dataset $\mathcal{T} = \{\boldsymbol{P}^t_i\}_{i = 1}^{N_t}$, however, contains only unlabeled point clouds.
Besides the trainable parameters, the model is additionally defined via its hyperparameters.
The anchors
$\Psi = \{ (\psi^{(x)}, \psi^{(y)}, \psi^{(z)}, \psi^{(w)}, \psi^{(l)}, \psi^{(h)}, \psi^{(\theta)}) \} \subset \Theta$
provide a starting point for the regression head, where the network predicts residuals relative to an anchor, instead of regressing the absolute values.
The anchors' width, length and height are usually handpicked to match the average size of annotated objects from $\mathcal{S}$.

The size discrepancy between objects in $\mathcal{S}$ and $\mathcal{T}$ induces an apparent domain gap, as illustrated in Figure~\ref{fig:density_width_length_oracle}.
With our \emph{unsupervised anchor calibration}, we adapt the anchor sizes to the target dataset $\mathcal{T}$ without any supervision.
As depicted in Figure~\ref{fig:method_overview}, using a model $H(\,\cdot\,|\,\Theta)$ pretrained on the source data, we first construct a reference feature database (Section~\ref{sec:ref_feat_db}) by accumulating the proposal feature vectors from the source domain $\mathcal{S}$.
Then, we iteratively perturb the model's anchor sizes by a small amount $\varepsilon$, while leaving \emph{all} other parameters unchanged
$(\psi^{(x)}, \psi^{(y)}, \psi^{(z)}, \psi^{(w)} + \varepsilon^{(w)}, \psi^{(l)} + \varepsilon^{(l)}, \psi^{(h)} + \varepsilon^{(h)}, \psi^{(\theta)})$
and, analogous to the reference feature database, compute a target feature database from $\mathcal{T}$.
The fitness of such a target feature database to the source feature database (Section~\ref{sec:target_fitness_quantification}) provides a stochastic gradient, which we utilize to adapt the model's anchor sizes (Section~\ref{sec:anchor_calibration}).
Our approach yields anchors which are specifically tailored to the given model for the target domain, without any retraining.

\subsection{Reference Feature Database}
\label{sec:ref_feat_db}

Using the model $H(\cdot\,|\,\Theta)$, pretrained on the source data $\mathcal{S}$, we first generate a reference feature database
\begin{equation}
  \mathcal{F}^s = \left\{
  RPN(
  \boldsymbol{P}^s \,|\, \Theta
  )
  \delta_{ [ \hat{c} > \tau ]}
  \;|\; \forall \boldsymbol{P}^s \in \mathcal{S}
  \right\}
  \text{.}
  \label{eq:ft_db}
\end{equation}
We select the $RPN$ features of the frame $\boldsymbol{P}^s$ where the respective prediction score $\hat{c}$, obtained from the classification head, exceeds a certain threshold $\tau$.
Ultimately, $\mathcal{F}^s$ is a latent feature database of the source database samples.

Depending on the size of the source dataset, $\mathcal{F}^s$ can have a large memory footprint.
Therefore, to compresses a potentially abundant reference feature database we fit a Gaussian Mixture Model (GMM) to the data.
A GMM consists of $K$ weighted Gaussian distributions, where the probability of observing a sample $f$ is defined as
\begin{equation}
  p(f\,|\,\Omega) = \sum_{i=1}^K \omega_i \cdot \mathcal{N} (f\,|\,\mu_i, \Sigma_i)
  ,\quad\text{where}
\end{equation}
\begin{equation}
  \mathcal{N}(f\,|\,\mu_i, \Sigma_i) =
  \frac{\exp \left( -\frac{1}{2} (f - \mu_i)^T \Sigma_i^{-1}(f - \mu_i) \right)}{\sqrt{(2 \pi)^K \lvert \Sigma_i \rvert}}
  \text{.}
\end{equation}
Here, $\omega_i$ weighs the $i^{\text{th}}$ Gaussian distribution, which is defined by its mean vector $\mu_i$ and covariance matrix $\Sigma_i$.
We fit the GMM parameters $\Omega_{\mathcal{F}^s} = \{ (\omega_i, \mu_i, \Sigma_i) \}_{i = 1}^K$ to a source feature database $\mathcal{F}^s$ using Expectation-Maximization~\cite{stat:em}.

\subsection{Target Fitness Quantification}
\label{sec:target_fitness_quantification}

The reference probability model $\Omega_{\mathcal{F}^s}$ describes ideal latent instances, \ie high-dimensional feature vectors that are abstractions of a complete object without background noise.
The inference on the target data, using the source model, will inevitably lead to predictions which include background noise or do not contain the object completely.
However, we can reduce such noisy predictions by selecting adequate anchor sizes. 
In this section, we show how we quantify the quality of the selected anchors.

Using the source model $H( \, \cdot \, | \, \Theta )$, we compute the target database $\mathcal{F}^t$ from the target dataset $\mathcal{T}$, analogous to Equation~\eqref{eq:ft_db}.
However, during this inference, we ignore the network's size residual predictions $(\Delta^{(w)}, \Delta^{(l)}, \Delta^{(h)})$ and use only matched anchor sizes, \ie $(\psi^{(x)} + \Delta^{(x)}, \psi^{(y)} +  \Delta^{(y)},  \psi^{(z)} + \Delta^{(z)}, \psi^{(w)}, \psi^{(l)}, \psi^{(h)}, \psi^{(\theta)} + \Delta^{(\theta)})$.
This isolates the influence of the selected anchors, because otherwise, the consecutive regression stage of the detector would obfuscate the changes.

To quantify the fitness of the anchors, we leverage the joint probability density
\begin{equation}
  p( \mathcal{F}^t\,|\,\Omega_{\mathcal{F}^s} ) = \prod_{f_i^t \in \mathcal{F}^t} p ( f_i^t\,|\,\Omega_{\mathcal{F}^s} )
  \text{,}
\end{equation}
where $\Omega_{\mathcal{F}^s}$ are the parameters of the reference probability model and the features $f_i^t$ are independent and identically distributed.
To avoid numerical instabilities we do not explicitly employ the joint probability but instead use the per-sample average log-likelihood
\begin{equation}
  \mathcal{L} ( \mathcal{F}^t \, | \,  \Omega_{\mathcal{F}^s}) =
  \frac{1}{ \lvert \mathcal{F}^t \rvert }
  \sum_{ f_i^t \in \mathcal{F}^t } \log p ( f_i^t\,|\,\Omega_{\mathcal{F}^s} )
  \text{.}
  \label{eq:fitness}
\end{equation}
The sum of the logarithms is numerically stable and the cardinality normalization ensures independence of the number of target features.

Smaller values of $\mathcal{L}$ indicate unfit predictions, \eg if target features include more background clutter or less object cues compared to the source features, and is maximized for anchors which are optimal for the target domain.
We present an experiment which demonstrates this behavior in Figure~\ref{fig:fitness_vs_map}.

\subsection{Anchor Calibration}
\label{sec:anchor_calibration}

Using the reference feature database $\mathcal{F}^s$ and fitness quantification $\mathcal{L}$, we can now compute the optimal anchors $\Psi^*$ for a model $H (\, \cdot \, | \, \Theta )$ by optimizing
\begin{equation}
  \Psi^* = 
  \argmin_{ \Psi }
  -\mathcal{L}( \mathcal{F}^t_{\Psi} \, | \, \Omega_{\mathcal{F}^s} )
  \text{,}
  \label{eq:objective}
\end{equation}
where $\Psi^* \subset \Theta$ represents the optimal anchors for the target dataset $\mathcal{T}$.
We outline our approach in Algorithm~\ref{alg:pseudo_code}.
\begin{algorithm}
\caption{Pseudocode of our SAILOR method}\label{alg:pseudo_code}
\begin{algorithmic}
  \Require 3D detection model $H( \, \cdot \, | \, \Theta )$ and anchors $\Psi$ optimized on a source dataset $\mathcal{S}$; Labeled source $\mathcal{S}$ and unlabeled target $\mathcal{T}$ dataset
  \Ensure Optimized anchors $\Psi^*$ for $\mathcal{T}$
  \State Generate $\mathcal{F}^s$ \Comment{Section~\ref{sec:ref_feat_db}}
  \State Fit GMM parameters $\Omega_{\mathcal{F}^s}$ to $\mathcal{F}^s$ with EM
  \State $\Psi^* = \Psi$
\While{Termination criteria is not reached}
  \State Randomly sample small $\varepsilon$
  \State $\Psi' = \Psi^* + \varepsilon$ \Comment{Anchor size perturbation}
  \State Generate $\mathcal{F}^t_{\Psi'}$ \Comment{Section~\ref{sec:target_fitness_quantification}}
  \State Update $\Psi^*$ with $\nabla \mathcal{L} ( \mathcal{F}^t_{\Psi'} \, | \, \Omega_{\mathcal{F}^s} )$ \Comment{Section~\ref{sec:anchor_calibration}}
  \EndWhile
\end{algorithmic}
\end{algorithm}

Since Equation~\eqref{eq:objective} is not differentiable \wrt the model hyperparameters, we do not use standard gradient methods.
Instead, we utilize Differential Evolution (DE)~\cite{optim:diff_evol}, a stochastic optimization technique.
At each iteration, given a population vector, DE constructs a mutation vector.
A trial candidate is constructed in a crossover phase by mixing the mutation vector with a candidate solution.
If the fitness of the trial candidate exceeds the current candidate solution, it becomes the next candidate solution.
This is repeated until a convergence criterion is met.

\begin{figure}[t]
  \begin{center}
    \includegraphics[width=\linewidth]{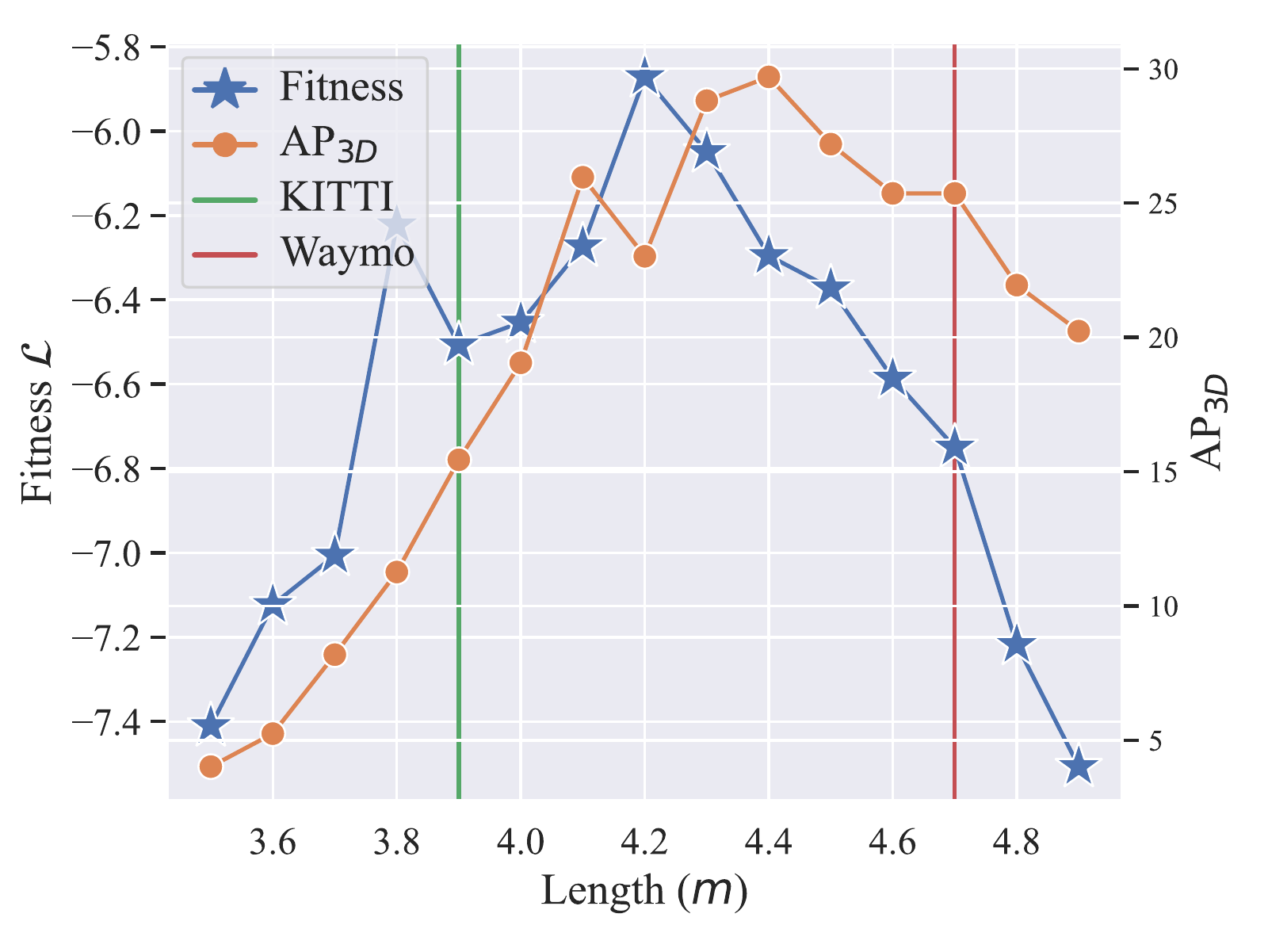}
  \end{center}
  \caption{
    Comparing anchor fitness (Equation~\eqref{eq:fitness}) and Average Precision.
    The model is trained on Waymo~\cite{3ddataset:waymo}, whereas the fitness and AP are assessed on KITTI~\cite{3ddataset:kitti}.
    Here, we fix width and height of the anchors and vary only the length.
    The fitness is computed without any annotation in the target domain and strongly correlates with the AP$_{3D}$.
    We denote ground truth Waymo and KITTI anchors in red and green, respectively.
  }
  \label{fig:fitness_vs_map}
\end{figure}

To accelerate the optimization, we first perform a linear sweep for each parameter separately.
Figure~\ref{fig:fitness_vs_map} depicts an example of this search.
We freeze other anchor sizes, and while varying a single parameter, assess the fitness at each step.
We use the parameter with the highest overall fitness as the initial candidate solution for the following joint optimization phase.
The reduced search space expedites the final joint fine-tuning.

Starting from this initial candidate solution $\Psi^*$, we jointly optimize the length, width and height.
For this, we generate the initial population vector by uniformly sampling $N_p$ anchor values
$\boldsymbol{\Psi} = \{ \Psi_i \}_{i=1}^{N_p}$
, where the sampling range is a percentage of the source anchor value.
A mutation vector is constructed as
\begin{equation}
  \Psi' = \Psi_{0} + \eta \cdot ( \Psi_{r_1} - \Psi_{r_2} )
  \text{,}
\end{equation}
where $\Psi_{0}$ is the population vector member with the best fitness score, $r_1$ and $r_2$ are two randomly selected indices, and $\eta$ is the mutation amplitude constant.
We terminate the optimization when the loss converges or when we reach a maximum number of iterations.

\section{Experiments}

\begin{table*}
  \begin{center}
    \resizebox{0.95\textwidth}{!}{
      \begin{tabular}{l c c c c}
        \toprule[1pt]
                                           &                       & \multicolumn{3}{c}{Target}                                                                                                                                                                                                                     \\
        \cmidrule(lr){3-5}
                                           &                       & \textbf{KITTI}                                                     & \textbf{Waymo}                                                                         & \textbf{nuScenes}                                                                \\
        \cmidrule(lr){3-5}
                                           &                       & AP$_{3D}$@R$11$                                                    & L$1$\_AP                                                                               & mAP                                                                              \\
        Source                             & Method                & Car / Pedestrian / Cyclist                                         & Vehicle / Pedestrian / Cyclist                                                         & Car / Pedestrian / Bicycle                                                       \\
        \cmidrule(lr){1-1}
        \cmidrule(lr){2-2}
        \cmidrule(lr){3-5}
        \multirow{6}{*}{\textbf{KITTI}}    & Source Anchors        & $\bestresult{78.59}$ / $\secbresult{54.58}$ / $\bestresult{64.53}$ & $\phantom{0}3.55$ / $\phantom{0}8.69$ / $\phantom{0}9.93$                              & $14.66$ / $\phantom{0}0.0\phantom{0}$ / $\phantom{0}0.0\phantom{0}$              \\
        \cmidrule(lr){2-5}
                                           & SN$^\dag$             & -                                                                  & $\phantom{0}2.80$ / $\phantom{0}2.05$ / $\phantom{0}1.65$                              & $\bestresult{20.33}$ / $\phantom{0}0.0\phantom{0}$ / $\phantom{0}0.0\phantom{0}$ \\
                                           & OT$^\dag$             & -                                                                  & $\phantom{0}6.51$ / $\bestresult{13.94}$ / $\bestresult{16.81}$                        & $14.65$  / $\phantom{0}0.0\phantom{0}$ / $\phantom{0}0.0\phantom{0}$             \\
                                           & ROS$^\dag$            & -                                                                  & $\phantom{0}3.32$ / $\phantom{0}2.75$ / $\phantom{0}2.29$                              & $\secbresult{18.54}$ / $\phantom{0}0.0\phantom{0}$ / $\phantom{0}0.0\phantom{0}$ \\
                                           & Target Anchors$^\dag$ & -                                                                  & $\bestresult{11.69}$ / $\secbresult{10.40}$ / $\secbresult{12.15}$                     & $15.88$ / $\phantom{0}0.0\phantom{0}$ / $\phantom{0}0.0\phantom{0}$              \\
        \cmidrule(lr){2-5}
                                           & Ours                  & $\secbresult{77.79}$ / $\bestresult{55.49}$ / $\secbresult{63.52}$ & $\phantom{0}\secbresult{8.26}$ / $\phantom{0}8.87$ / $\phantom{0}9.52$                 & $16.12$ / $\phantom{0}0.0\phantom{0}$ / $\phantom{0}0.0\phantom{0}$              \\
        \midrule
        \multirow{6}{*}{\textbf{Waymo}}    & Source Anchors        & $23.94$ / $59.99$ / $52.32$                                        & $\bestresult{74.84}$ / $\bestresult{71.59}$ / $\bestresult{66.49}$                     & $32.45$ / $17.99$ / $\phantom{0}0.0\phantom{0}$                                  \\
        \cmidrule(lr){2-5}
                                           & SN$^\dag$             & $24.11$ / $\bestresult{63.51}$ / $52.83$                           & -                                                                                      & $34.05$ / $17.24$ / $\phantom{0}0.0\phantom{0}$                                  \\
                                           & OT$^\dag$             & $38.79$ / $53.31$ / $\bestresult{61.82}$                           & -                                                                                      & $\bestresult{37.40}$ / $\secbresult{19.99}$ / $\phantom{0}0.0\phantom{0}$        \\
                                           & ROS$^\dag$            & $\secbresult{43.00}$ / $\secbresult{62.39}$ / $51.09$              & -                                                                                      & $35.91$ / $18.62$ / $\phantom{0}0.0\phantom{0}$                                  \\
                                           & Target Anchors$^\dag$ & $39.69$ / $49.96$ / $\secbresult{55.45}$                           & -                                                                                      & $\secbresult{37.30}$ / $\bestresult{20.32}$ / $\phantom{0}0.0\phantom{0}$        \\
        \cmidrule(lr){2-5}
                                           & Ours                  & $\bestresult{58.02}$ / $61.60$ / $53.04$                           & $\secbresult{73.98}$ / $\secbresult{69.48}$ / $\secbresult{66.19}$                     & $32.34$ / $17.27$ / $\phantom{0}0.0\phantom{0}$                                  \\
        \midrule
        \multirow{6}{*}{\textbf{nuScenes}} & Source Anchors        & $26.37$ / $10.37$ / $\secbresult{22.90}$                           & $29.59$ / $\phantom{0}2.99$ / $\phantom{0}2.66$                                        & $\bestresult{59.93}$ / $\bestresult{11.03}$ / $\bestresult{\phantom{0}0.23}$     \\
        \cmidrule(lr){2-5}
                                           & SN$^\dag$             & $33.67$ / $\bestresult{31.67}$ / $14.29$                           & $\bestresult{34.48}$ / $\phantom{0}2.34$ / $\phantom{0}\secbresult{4.08}$              & -                                                                                \\
                                           & OT$^\dag$             & $36.42$ / $18.15$ / $\phantom{0}8.84$                              & $31.37$ / $\bestresult{\phantom{0}8.22}$ / $\phantom{0}0.13$                           & -                                                                                \\
                                           & ROS$^\dag$            & $\secbresult{56.28}$ / $\secbresult{26.88}$ / $19.76$              & $\secbresult{32.10}$ / $\phantom{0}\secbresult{7.73}$ / $\bestresult{\phantom{0}7.03}$ & -                                                                                \\
                                           & Target Anchors$^\dag$ & $\bestresult{57.93}$ / $18.57$ / $11.68$                           & $31.70$ / $\phantom{0}6.69$ / $\phantom{0}0.41$                                        & -                                                                                \\
        \cmidrule(lr){2-5}
                                           & Ours                  & $55.10$ / $\phantom{0}4.64$ / $\bestresult{26.37}$                 & $29.00$ / $\phantom{0}2.07$ / $\phantom{0}2.81$                                        & $\secbresult{59.41}$ / $\secbresult{10.98}$ / $\phantom{0}\secbresult{0.12}$     \\
        \bottomrule[1pt]
      \end{tabular}
    }
  \end{center}
  \caption{
    Extensive comparison of Statistical Normalization (SN)~\cite{3dda:train_germany}, Output Transformation (OT)~\cite{3dda:train_germany} and Random Object Scaling (ROS)~\cite{3dda:st3d}
    with our method on popular autonomous driving datasets KITTI~\cite{3ddataset:kitti}, Waymo~\cite{3ddataset:waymo} and nuScenes~\cite{3ddataset:nuscenes}.
    We use the whole KITTI and nuScenes dataset for both training and evaluation and utilize 20\% of Waymo v$1.2$ for training and full evaluation dataset.
    The model is trained on the source data using the configuration as in OpenPCDet~\cite{github:openpcdet}.
    We always evaluate the \emph{last checkpoint} using the target dataset evaluation pipeline.
    For KITTI, we show moderate case of AP$_{3D}$@R$11$ for Car / Pedestrian / Cyclist thresholded at $0.7$, $0.5$ and $0.5$, respectively.
    We report L$1$\_AP for Vehicle / Pedestrian / Cyclist on Waymo, whereas for nuScenes, we provide mAP for the classes Car / Pedestrian / Bicycle.
    We mark weakly-supervised approaches with $^\dag$, since they require target object statistics.
  }
  \label{table:main_eval}
\end{table*}

We focus our evaluation on KITTI~\cite{3ddataset:kitti}, Waymo~\cite{3ddataset:waymo} and nuScenes~\cite{3ddataset:nuscenes}, and include Lyft~\cite{3ddataset:lyft} for further generalization evaluations.
Inter-domain anchor calibration demonstrates the capability, whereas intra-domain calibration verifies the correctness of our approach.
The variety of the data, \eg sparse to dense (nuScenes~$\leftrightarrow$~Waymo) or a large size gap (KITTI~$\leftrightarrow$~Waymo), faithfully represents an arbitrary use-case.
The source and target data are the training and validation splits of the datasets mentioned above, respectively.
We always perform the evaluation following the evaluation protocol of the target dataset.

\subsection{Comparison with the State-of-the-Art}

To the best of our knowledge, there is no existing work in the unsupervised domain to compare to.
Therefore, we compare SAILOR\footnote{\url{https://github.com/malicd/sailor}} with the widely adapted weakly-supervised approaches that address the object size bias, \ie Statistical Normalization (SN)~\cite{3dda:train_germany}, Output Transformation (OT)~\cite{3dda:train_germany} and Random Object Scaling (ROS)~\cite{3dda:st3d}.

For SN, we follow the original publication and set the model anchors to the average of the source and target domain and scale the labeled source bounding boxes and points inside using the difference between the statistics.
Similarly, ROS exploits this knowledge, but in a more coarse way.
Knowing that KITTI objects are smaller than Waymo's, we can apply the appropriate scaling for ROS during the adaptation.
We retrain both models according to the source configuration.
OT does not require training but instead uses the source model and directly adds the difference to the predictions.
We refer the reader to the supplementary material, where we list the exact anchor configuration for reproducibility.

\begin{figure*}
     \centering
     \begin{subfigure}[b]{0.24\textwidth}
         \centering
         \includegraphics[width=\textwidth]{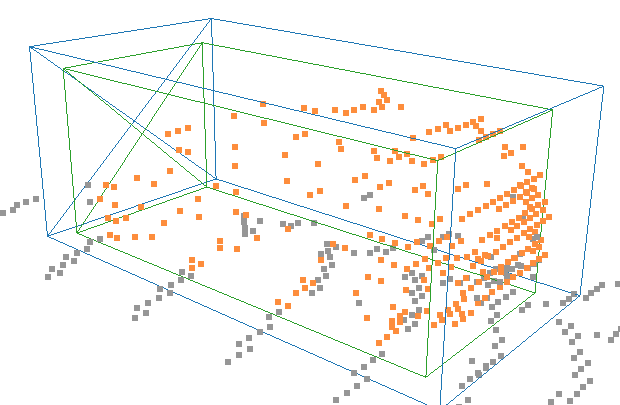}
         \caption{Source only}
         \label{fig:qualitative_source}
     \end{subfigure}
     \hfill
     \begin{subfigure}[b]{0.24\textwidth}
         \centering
         \includegraphics[width=\textwidth]{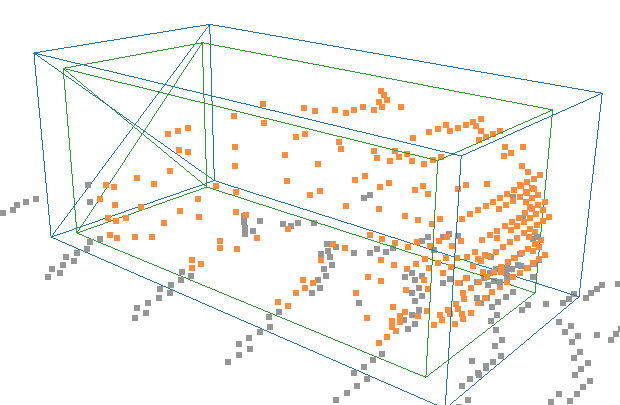}
         \caption{SN}
         \label{fig:qualitative_sn}
     \end{subfigure}
     \hfill
     \begin{subfigure}[b]{0.24\textwidth}
         \centering
         \includegraphics[width=\textwidth]{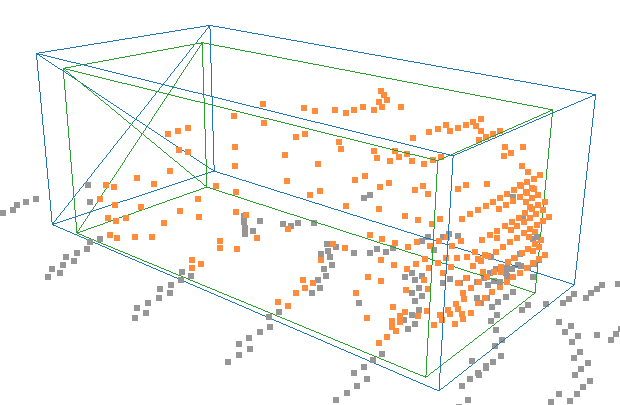}
         \caption{ROS}
         \label{fig:qualitative_ros}
     \end{subfigure}
     \hfill
     \begin{subfigure}[b]{0.24\textwidth}
         \centering
         \includegraphics[width=\textwidth]{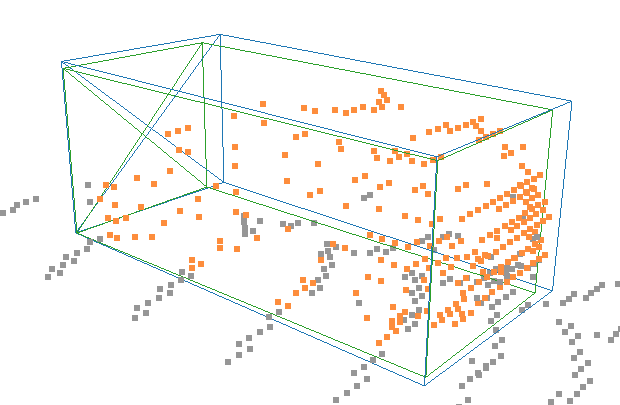}
         \caption{Ours}
         \label{fig:qualitative_ours}
     \end{subfigure}
        \caption{
          Qualitative comparison of source only, Statistical Normalization (SN)~\cite{3dda:train_germany}, Random Object Scaling (ROS)~\cite{3dda:st3d} and our method on Waymo $\rightarrow$ KITTI case.
          We indicate the ground truth box in green and the predicted boxes in blue.
          The object points, according to the ground truth annotation, are shown in orange.
          Best viewed on screen.
        }
        \label{fig:quantitative}
\end{figure*}

During the evaluation, we directly employ the anchors optimized by SAILOR.
We do not perform any retraining whatsoever.
Table~\ref{table:main_eval} demonstrates the potential of our method, given a \partatwo~source model.
In the case of Waymo~$\rightarrow$~KITTI, we observe an improvement of $34$ AP$_{3D}$@R$11$ for the Car class, beating even the weakly-supervised approaches by a large margin.
We depict this improvement in Figure~\ref{fig:quantitative}.
Pedestrian and Cyclists are mostly unchanged, except for a slight improvement, mainly due to having similar anchors.
We report similar behavior in the nuScenes $\rightarrow$ KITTI experiment, where we observe an improvement of almost $28$ AP$_{3D}$@R$11$.
We also note that the slight precision increase for the Cyclists class stems from the significant height difference (due to different labeling policies) between the two datasets.

In situations where a source model shows inferior performance and the size domain gap is large, \eg KITTI~$\rightarrow$~Waymo and KITTI~$\rightarrow$~nuScenes, we still accomplish substantial relative gain.
Calibrating KITTI anchors on Waymo, we achieve a relative improvement over 200\%.
This tremendous increase is significantly better than the weakly-supervised SN and ROS, which even degrade the performance slightly.
Note that the KITTI~$\rightarrow$~Waymo scenario is well-known to be extremely challenging and has thus been often neglected from evaluations, with very few exceptions, \eg~\cite{3dda:fast3d}.
The overall low performance on KITTI~$\rightarrow$~nuScenes stems from a small source dataset and the sparsity of the nuScenes point clouds, where the detector struggles.
Thus, there are only very few objects with insufficient latent semantics to apply our method.

The datasets that share similar anchors, \eg Waymo~$\leftrightarrow$~nuScenes, do not exhibit a substantial change in the overall evaluation score.
Since SAILOR introduces no additional hyperparameters, the weakly-supervised approaches perform favorable in this scenario, as they can exploit the target domain statistics.
However, note that similar improvements can also be achieved by fiercer augmentation, as shown in~\cite{3dda:quantifying_augmentation}.

Similarly, when our source and target data come from the same dataset, \ie the diagonal in Table~\ref{table:main_eval}, we report an insignificant change in the results.
We performed this additional experiment mainly as a sanity check.

\begin{table}
  \begin{center}
    \resizebox{0.95\columnwidth}{!}{
      \begin{tabular}{c c c}
        \toprule
        Task                                       & Method                & AP$_{3D}$@R$11$                                                  \\
        \cmidrule(lr){1-1} \cmidrule(lr){2-2} \cmidrule(lr){3-3}
        \multirowcell{6}{Waymo $\rightarrow$ Lyft} & Source Anchors        & $49.88$ / $37.17$    / $25.62$                                   \\
        \cmidrule(lr){2-3}
                                                   & SN$^\dag$             & $44.96$ / $33.98$    / $17.00$                                   \\
                                                   & OT$^\dag$             & $\bestresult{51.58}$ / $37.96$    / $15.32$                      \\
                                                   & ROS$^\dag$            & $49.39$ / $34.06$    / $25.07$                                   \\
                                                   & Target Anchors$^\dag$ & $51.57$ / $37.02$    / $16.15$                                   \\
        \cmidrule(lr){2-3}
                                                   & Ours                  & $49.31$     / $\bestresult{38.27}$        / $\bestresult{26.01}$ \\
        \midrule
        \multirowcell{6}{KITTI $\rightarrow$ Lyft} & Source Anchors        & $20.61$ / $13.94$    / $14.24$                                   \\
        \cmidrule(lr){2-3}
                                                   & SN$^\dag$             & $\bestresult{28.27}$ / $13.36$    / $10.81$                      \\
                                                   & OT$^\dag$             & $20.83$ / $16.83$    / $10.40$                                   \\
                                                   & ROS$^\dag$            & $25.86$ / $16.33$    / $\phantom{0}8.05$                         \\
                                                   & Target Anchors$^\dag$ & $26.12$ / $17.24$    / $\bestresult{17.24}$                      \\
        \cmidrule(lr){2-3}
                                                   & Ours                  & $26.48$     / $\bestresult{17.27}$        / $14.20$              \\
        \bottomrule
      \end{tabular}
    }
  \end{center}
  \caption{
    Results of the adaptation tasks to Lyft.
    We report AP$_{3D}$@R$11$ for the classes Car / Pedestrian / Bicycle.
  }
  \label{table:to_lyft}
\end{table}

To further demonstrate the generalization capabilities of our method, we perform additional experiments on the Lyft~\cite{3ddataset:lyft} dataset.
Our findings, which we summarize in Table~\ref{table:to_lyft}, confirm our initial experiments.
In cases where the source and target anchors are similar, \eg Waymo $\rightarrow$ Lyft, we report slight performance gains.
Additionally, when the object size gap is large, \eg KITTI~$\rightarrow$~Lyft for class car, we report increase of around 6 AP points.
Similarly to KITTI~$\rightarrow$~nuScenes, we found that further improvement is limited by an ineffective source model caused by a small source dataset.
However, due to sufficient density of the Lyft point clouds (LiDAR with 64 beams), we do not observe any precision drop.

Moreover, we demonstrate that combining our method with existing weakly-supervised approaches leads to competitive results in a completely unsupervised manner.
For this, we replace the target statistics required by SN and ROS by the anchor sizes calibrated with SAILOR.
The results in Table~\ref{table:sn_ros_with_sailor} show that this is well suited to make ROS fully unsupervised, whereas for SN it works well for vehicles, but is not suitable for pedestrians and cyclists.
This issue is due to SN, which is tailored explicitly to knowing the target statistics and requires additional manual fine-tuning (which we omitted due to the unsupervised setting of this experiment) to achieve the best results.
Note that we are not interested in estimating the actual target domain object sizes, but in estimating the optimal target domain anchors for the given source model.
Using these calibrated anchors to guide ROS, our unsupervised ROS variant performs on par with the weakly-supervised one, \eg Waymo $\rightarrow$ KITTI, or even better, \eg nuScenes $\rightarrow$ KITTI.
\begin{table}
  \begin{center}
    \resizebox{\columnwidth}{!}{
      \begin{tabular}{ccc}
        \toprule[0.8pt]
        Task                                          & Method        & AP$_{3D}$@R$11$                                                    \\
        \cmidrule(lr){1-1}
        \cmidrule(lr){2-2}
        \cmidrule(lr){3-3}
        \multirow{4}{*}{Waymo $\rightarrow$ KITTI}    & SN$^\dag$     & $24.11$ / $\bestresult{63.51}$ / $\bestresult{52.83}$              \\
                                                      & ROS$^\dag$    & $\bestresult{43.00}$ / $62.39$ / $51.09$                           \\
        \cmidrule(lr){2-3}
                                                      & SN w/ SAILOR  & $22.31$ / $55.68$ / $38.96$                                        \\
                                                      & ROS w/ SAILOR & $39.82$ / $60.40$ / $48.61$                                        \\
        \midrule
        \multirow{4}{*}{nuScenes $\rightarrow$ KITTI} & SN$^\dag$     & $33.67$ / $31.67$ / $14.29$                                        \\
                                                      & ROS$^\dag$    & $56.28$ / $26.88$ / $19.76$                                        \\
        \cmidrule(lr){2-3}
                                                      & SN w/ SAILOR  & $39.44$ / $\phantom{0}0.51$ / $\phantom{0}0.00$                    \\
                                                      & ROS w/ SAILOR & $\bestresult{57.98}$ / $\bestresult{34.59}$ / $\bestresult{24.15}$ \\
        \bottomrule[0.8pt]
      \end{tabular}
    }
  \end{center}
  \caption{
    SN and ROS become entirely unsupervised in combination with SAILOR, otherwise they are weakly-supervised and denoted with $^\dag$.
    The results are the moderate case for Car / Pedestrian / Cyclist classes.
  }
  \label{table:sn_ros_with_sailor}
\end{table}

\subsection{Ablation Studies}
We conduct our ablation experiments to investigate the interaction between the components of our system, as well as to study the vital points of our pipeline.
We also indicate the known caveats and demonstrate how we overcome them.

\paragraph{Joint Optimization}
We demonstrate the benefits of our joint optimization in Table~\ref{table:ablation_linear_gd}, which shows that linear search already provides a decent performance.
However, it optimizes each component individually, disregarding their entanglement.
We leverage this even further to boost the performance using joint optimization from Section~\ref{sec:anchor_calibration}.

In some instances, \eg nuScenes $\rightarrow$ KITTI for the class Pedestrian, DE degrades the result.
Our observation shows that this happens for two reasons.
When the number of instances in the source domain is too low for the proper estimation of the GMM parameters, \eg KITTI~$\rightarrow$~Waymo for Pedestrians and Cyclists, our objective function becomes harder to optimize due to the appearing local minima.
A similar effect can be observed by inadequate latent representations of the source model, caused by the point cloud sparsity, \eg nuScenes $\rightarrow$ KITTI for Pedestrians.
\begin{table}
  \begin{center}
    \resizebox{\columnwidth}{!}{
      \begin{tabular}{clc}
        \toprule
        Task                                          & Method  & AP$_{3D}$@R$11$                                                    \\
        \cmidrule(lr){1-1}
        \cmidrule(lr){2-2}
        \cmidrule(lr){3-3}
        \multirow{3}{*}{Waymo $\rightarrow$ KITTI}    & Source  & $23.94$ / $59.99$ / $52.32$                                        \\
                                                      & LS      & $50.57$ / $58.27$ / $51.49$                                        \\
                                                      & LS + DE & $\bestresult{58.02}$ / $\bestresult{61.59}$ / $\bestresult{53.04}$ \\
        \midrule
        \multirow{3}{*}{nuScenes $\rightarrow$ KITTI} & Source  & $26.37$ / $10.37$ / $22.90$                                        \\
                                                      & LS      & $36.85$ / $\bestresult{10.61}$ / $25.35$                           \\
                                                      & LS + DE & $\bestresult{55.10}$ / $\phantom{0}4.64$ / $\bestresult{26.37}$    \\
        \bottomrule
      \end{tabular}
    }
  \end{center}
  \caption{
    Detection performance \wrt the optimization strategy for the classes Car / Pedestrian / Cyclist.
    \textbf{LS} stands for Linear Search and \textbf{DE} is Differential Evolution.
  }
  \label{table:ablation_linear_gd}
\end{table}

\begin{figure}[t]
  \begin{center}
    \includegraphics[width=0.75\linewidth]{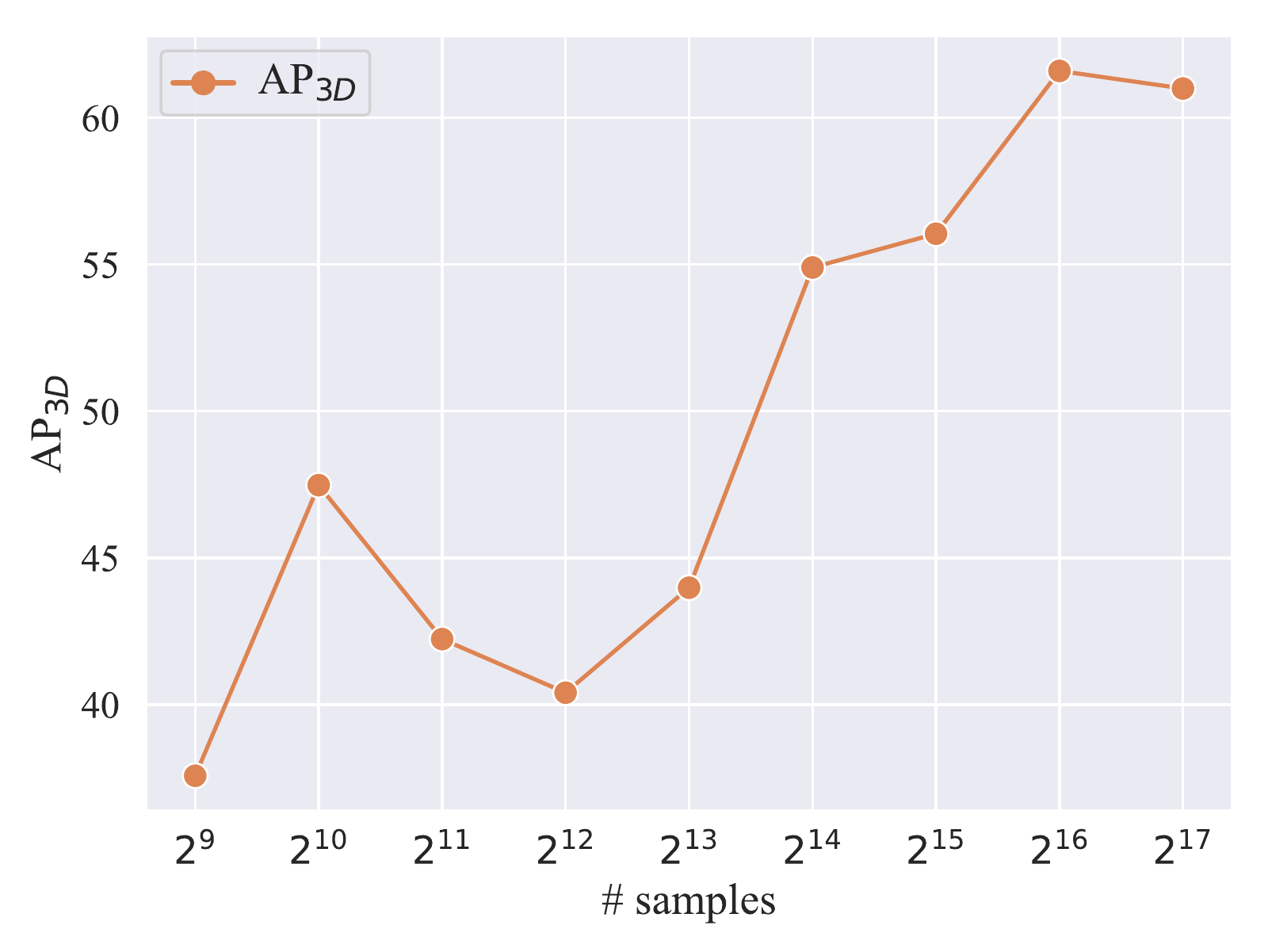}
  \end{center}
  \caption{
    Applying SAILOR on Waymo $\rightarrow$ KITTI while varying the number of instances used for GMM fitting.
    We report Average Precision for the moderate case of cars.
  }
  \label{fig:n_samples_vs_map}
\end{figure}

\paragraph{Smoothness of the Optimization Objective}
As previously hinted, the number of samples used for the GMM fitting plays an important role in our method.
Our objective function from Equation~\eqref{eq:fitness} gets smoother with the number of instances we use.
If we build the source feature database from underrepresented samples, it becomes hard to optimize many appearing local minima.
This implicitly affects the performance on the target dataset, as depicted in Figure~\ref{fig:n_samples_vs_map}.
On the other hand, SAILOR is not sensitive to the number of components in the GMM at all.
We found that any reasonable choice of $K \geq 4$ works similarly well.

\section{Conclusion}
We presented SAILOR, an unsupervised approach for anchor calibration on the target domain.
We estimate an optimal anchor configuration under the source model without prior knowledge.
We compare our approach with weakly-supervised methods, which are widely used for unsupervised 3D domain adaptation.
Moreover, SAILOR can be used as a stand-alone method or can make these weakly-supervised approaches completely unsupervised.

In this work, we focus explicitly on optimizing anchor sizes due to the tremendous domain gap across datasets, particularly KITTI $\leftrightarrow$ Waymo.
Note, however, that any model hyperparameter is optimizable with our SAILOR schema, even the checkpoint selection procedure, which we want to investigate in future experiments.

\paragraph{Acknowledgments}
We gratefully acknowledge the financial support by the Austrian Federal Ministry for Digital and Economic Affairs, the National Foundation for Research, Technology and Development and the Christian Doppler Research Association.
The presented experiments have been achieved (in part) using the Vienna Scientific Cluster.

{\small
\bibliographystyle{ieee_fullname}
\bibliography{abbrv_short,egbib}
}

\end{document}